# Towards Transparent Mental Health Insights: An Explainable AI Model for Career-Related Depression and Anxiety Among University Students Using Structured Data

**Arsham Azam [1,2], Rasikh Ali [1,2], Tayyaba Farhat [1,2]* and Sheeraz Akram [1,2,3]**

[1] Faculty of Computer Science and Information Technology, The Superior University, Lahore 54600, Pakistan; rasikhali1234@gmail.com; arshamrana36@gmail.com; tayyaba.farhat@superior.edu.pk
[2] Intelligent Data Visual Computing Research (IDVCR), Lahore 54600, Pakistan;
[3] Information Systems Department, College of Computer and Information Sciences, Imam Mohammad Ibn Saud Islamic University (IMSIU), Riyadh 12571, Saudi Arabia; sAkram@imamu.edu.sa
* Correspondence: tayyaba.farhat@superior.edu.pk;

**Abstract:** The growing incidence of career anxiety and depression among university students presents a vital challenge to mental health and academic achievement. This study suggests an Explainable AI (XAI) led framework using multimodal data and Federated Learning (FL) to identify early indicators of career-related mental health problems in a privacy-preserving and culturally responsive way. The framework combines structured behavioral data, facial emotion feature from interview videos (e.g., head pose, gaze directions, and action unit intensity), and app usage patterns via an intermediate fusion neural network architecture with attention mechanisms. Label smoothing was adopted to improve model generalizability. In order to ensure data privacy, FL was used across multiple institutions so that collaborative training could be done without raw data sharing. Experimental evaluation was conducted using the Student Mental Health Survey dataset, which comprises structured responses from university students across Pakistan. The dataset includes variables such as age, gender, academic year, GPA, and self-reported indicators of mental health challenges, including depression and anxiety. Participants represent a diverse mix of undergraduate, graduate, and postgraduate students, providing a comprehensive overview of student mental health across different academic levels. Our model attained an F1-score of 89.12%, recall of 86.54%, accuracy of 92.08%, and precision of 91.88%. With Integrated Gradients and SHAP values, the model identified key behavioral markers of depression such as avoidance of direct gaze, lower facial expressiveness, frowning, and social withdrawal signs consistent with psychological theory. This research illustrates an interpretable, scalable, and context sensitive AI system for mental health pre-diagnosis with possible integration into student support services globally. Future studies will investigate cross cultural generalizability and real-time intervention to enhance early awareness of mental health and individualized care in the academic setting.

## 1. Introduction

In the past few years, depression and anxiety have been on the rise among university students globally, with career uncertainty and academic stress being major predisposing factors. Career depression and anxiety, which remain poorly identified within conventional diagnostic practices, can dramatically impact students' academic achievement, social functioning, and future planning. The problem is compounded by limited timely mental health care access and stigma related to seeking help. In such a scenario, Artificial Intelligence (AI) has promising potential to aid early detection and intervention. Conventional diagnostic techniques are based primarily on subjective assessments during clinical interviews, wherein mental health practitioners make observations of verbal and non-verbal responses like tone of voice, facial expressions, and body language. These techniques are resource-hungry and usually out of reach for most students, particularly in low-resource environments. To overcome these challenges, this research presents an Explainable AI (XAI) model that combines multimodal data such as facial behavior, structured academic and psychological data, and behavioral patterns to identify career-related depression and anxiety among university students. To maintain student anonymity and facilitate massive-scale collaborative learning, a Federated Learning (FL) setup is utilized. This keeps the data decentralized yet facilitates contribution towards training the overall model. In addition, XAI methods like SHAP and Integrated Gradients are used in the study to promote transparency so that mental health practitioners can identify the reasoning behind predictions made by the model.

This research aims not just to enhance detection performance via multimodal fusion and sophisticated deep learning architectures, but also to offer culturally and contextually informed insights geared to university students from varied educational systems. By facilitating explainable, privacy-preserving, and scalable mental health screening, this work supports the increasing convergence of AI and international student well-being.

## 2. Literature Review

Recent research has brought to light the expanding ability of ML in identifying mental illnesses like depression, anxiety, and career anxiety, most notably among college students. Artificial Neural Networks (ANNs) have been demonstrated to have high prediction accuracy, with some research achieving as high as 91%, surpassing conventional algorithms such as Random Forest, Support Vector Machine, and Naive Bayes, which plays a huge role in early intervention strategies that seek to assist students' mental well-being. Nonetheless, ethical issues like data privacy, bias in algorithms, and excessive dependency on autonomous systems need to be taken into account with great caution during the rollout of such models [1]. More and more, the ethical and practical value of explainable AI (XAI) is realized, with research supporting transparency within AI systems to encourage clinician trust and improved readability of model outputs. For example, one study of a computerized mental health program with 970 participants discovered Random Forest to have the highest test accuracy at 0.71, with the incorporation of XAI techniques such as SHAP and LIME contributing significantly toward model transparency and enhancing clinician decision-making [2]. Even with these developments, issues continue to exist in managing missing data, attaining high predictive accuracy, and maintaining personalized, ethical care [2]. Additionally, the psychological impact of repeated lockdowns has been researched to highlight the sophistication of detecting mental health, especially among children, and the necessity for multimodal methods that incorporate temporal dynamics and social context, typically performed with deep learning models like RNNs, CNNs, and transformers [3]. Research like Ntakolia et al. (2023) demonstrate that XAI methods, such as SHAP, may assist in the identification of prime contributing factors to mental health decline, although they also mention the necessity for more specified and representative research among vulnerable populations [3]. Additionally, Chadaga et al. (2023) and Kelsen et al. (2023) illustrated the usability of explainable AI models for diagnosing medical conditions such as Social Anxiety Disorder and depression through linguistic and text-based assessment and highlighted the use of understandable and personalized models [2,4]. In spite of sample size and data variety limitations, the use of XAI and deep learning in such systems has been viewed as a promising direction for enhancing the accuracy and reliability of mental health detection [2,4]. Also, research like that done by Hasan et al. (2023) and Yao et al. (2023) highlights the value of integrating multiple data sources, e.g., social media as well as physiological data, to improve early identification of mental illnesses and overcome dataset bias and computational difficulties [4,5]. A significant advancement in this area is the development of multi-modal systems, such as the one proposed by Wang (2023), which combines large language models with explainability frameworks to improve depression detection accuracy and explainability [6]. These technologies are being applied with growing relevance, as evidenced by Mahayossanunt et al. (2024) and Rajawat et al. (2024), where they were working on applying facial expression recognition and hybrid models to sustain high accuracy and interpretability in detecting depression and anxiety and laying the groundwork for real-time monitoring and early intervention [7,8]. In addition, research by Karimian (2024) and Kumar Saha et al. (2024) also emphasizes the potential of ML to revolutionize diagnosis in the field of mental health, with a call for better multimodal data fusion and incorporating these systems within clinical practice so that individualized and real-time care may be delivered [9,10]. Lastly, recent studies have demonstrated potential in stress recognition and satisfaction with life prediction using ML and deep neural networks, highlighting the wide range of applications of AI in mental health understanding and management [11,12]. Yet, constraints like dataset specificity, cultural homogeneity, and privacy issues persist, highlighting the necessity for ongoing innovation and interdisciplinarity to create ethical, interpretable, and effective AI solutions for mental health care [13].

**Table** *1.* Related works of depression prediction

| **Reference** | **limitation** | **Contribution** | **Explainable AI model** | **Test Accuracy** | **Algorithm** |
|---|---|---|---|---|---|
| [1] | Class imbalance, unclear career anxiety concept | Studied career anxiety, exam system stress | -- | 91% | Artificial Neural Network (ANN) Random Forest (RF) Support Vector Machine (SVM) Naive Bayes (NB) Decision Tree (DT) |
| [2] | Missing data, no feature selection impact | Proposed XAI to enhance transparency, trust | Explainable AI-Based Decision Support System | 60% | Random Forest (RF) classifier - Support Vector |
| | | | (XAI-DSS) | | Machine (SVM) Logistic Regression Naive Bayes |
| [3] | Limited data diversity, lack of explainability | Surveying AI in mental health via social media, highlighting XAI needs | LIME, SHAP, LRP | -- | Recurrent Neural Networks (RNNs) Long Short-Term Memory (LSTMs) Gated Recurrent Units (GRUs) Convolutional Neural Networks (CNNs) BERT (Bidirectional Encoder Representations from Transformers) Graph Neural Networks (GNNs) |
| [4] | Parent-report bias, COVID stress omission, replicability concerns, selection bias, generalizability issues. | comprehensive dataset, XAI pipeline for mood state analysis. | SHAP | 76% | Linear Regression (LR) Logistic Regression (LogR) Multi-Layer Perceptron (MLP) LightGBM Support Vector |
| | | | | | Machine (SVM) |
| [5] | Small sample size, limited to supervised learning, absence of deep learning, restricted symptom data. | DSM-5 dataset, XAI interpretation, statistical analysis | SHAP, LIME, Eli5, and QLattice. | 88% | Logistic Regression |
| [9] | Limited sociodemographic data, single platform focus, binary classification | Binary classification | BiLSTM<br>LIME | -- | Pretrained Language Model (PLM) |
| [14] | Technical complexity, data interoperability, computational demands | ML/DL analysis for mental health prediction, disease progression modeling | SHAP<br>LIME | -- | Support Vector Machines (SVM)<br>Decision Trees<br>Logistic Regression<br>Naive Bayes<br>Random Forest (RF)<br>Extreme Gradient Boosting (XGBoost)<br>Deep Learning Methods |
| [6] | Small sample size of only 6 concussion clinicians | Identifying clinician challenges, deriving AI-RPM design considerations | -- | -- | -- |
| [7] | Small sample size, limited to English, single evaluation metric | Explainable depression detection, prompt-based LLMs for depression | Explainable Boosting Machine (EBM) | -- | Explainable Boosting Machine (EBM)<br>Decision Tree |
| | | prediction | | | (DT)<br>Logistic Regression (LR)<br>Llama-2-13b-chat<br>SUS-Chat-34B<br>Neural-chat-7b-v3-1 |
| [8] | Limited generalizability, demographic biases, high computational cost | Data analysis of depression patterns, ML/DL models with optimization | SHAP, LIME | 91.1% | Logistic Regression<br>Decision Tree<br>Random Forest<br>K-Nearest Neighbors (KNN)<br>Multi-Layer Perceptron (MLP)<br>Convolutional Neural Network (CNN) |
| [9] | Comprehensive explanations, multi-label classification, outperformance of non-fine-tuned ChatGPT model. | Multi-label framework for anxiety and depression, LLMs for feature extraction | -- | 90% | Naive Bayes<br>Decision Tree<br>Random Forest<br>Large Language Models (LLMs)<br>Explainability Dashboard |
| [10] | Data quality issues, model interpretability challenges | -- | -- | 99% | Neural networks<br>Supervised learning<br>Logistic regression |
| | | | | | Decision trees K-nearest neighbors (KNN) |
| [11] | Privacy restrictions, language barriers, limited model usability and enhancement. | Comparison of fusion models, Integrated Gradient for explainability. | Window block LSTM , Bi-LSTM, Integrated gradient | 91.67% | Transformer model |
| [12] | -- | Proposed Fusion Fuzzy Logic (FFL) model with CNNs | Fusion Fuzzy Logic (FFL) | 94.3% | Fusion Fuzzy Logic (FFL) with Deep Learning |
| [13] | Ignored temporal features, high computational cost, redundant raw image input. | Facial Motion Modeling Module (BiGRU), Facial Landmark Calibration Module (FLCM) | BiGRU (Facial Motion Modeling Module) | -- | Sparse Optical Flow & Pyramid Lucas-Kanade (PLK) |
| [15] | Overreliance on BDC scale, lack of biological factors, small dataset. | DeprMVM ensemble model, SVM-MLP layered framework | SHAP, LIME | 99.39% | RFC, KNN, SVM, XGB, MLP |
| [16] | Self-reported stress, EEG equipment limitations, dataset size, low generalizability. | QuadTPat feature extraction, CWNCA feature selection | QuadTPat-based Explainable Feature Engineering (XFE) model. | 92.94% | QuadTPat, CWNCA, tkNN, DLob. |
| [17] | Limited generalizability (specific context), lack of temporal dynamics. | Determinants of life satisfaction, High-accuracy ML models | AdaBoost, XGBoost, BERT, BioBERT, ClinicalBERT, COReBERT | 93.5%. | Random Forest (RF) Decision Tree - |

### *2.1. Gap Analysis*

In spite of the major breakthroughs in the use of machine learning (ML), deep learning (DL), and explainable artificial intelligence (XAI) for mental health detection, there remain major research gaps that limit the creation of robust, scalable, and culturally sensitive systems. One significant limitation is the poor fusion of multimodal information; though there are extensive studies on text, speech, or facial expressions independently, there are few that combine these modes to take advantage of their complementary strengths towards more accurate and context-conscious diagnosis. In addition, most of the current research focuses on Western populations and English-language data, which results in a lack of cultural and linguistic flexibility, especially in countries such as Pakistan where mental health symptom expressions may vary dramatically. Career anxiety, academic stress, and the impact of varying educational systems (e.g., semester vs. annual) are also significantly under researched, even though they are highly relevant among young adults in South Asia. A majority of existing models depend on static data sets and do not account for temporal dynamics, which cannot be used to detect the development or changes in mental health disorders over time. Moreover, although some studies utilize XAI methods like SHAP or LIME, their interpretability is also low, particularly for deep neural models such as GNNs and transformers, and few models are clinically validated or designed with assistance from mental health professionals. The prevalent use of minute, unbalanced, and non-standardized datasets also limits generalizability and real-world usability. Ethical issues regarding privacy, security of data, as well as the absence of clear data governance processes are also largely ignored, present serious challenges in terms of deployment in sensitive areas such as mental health. Lastly, although conceptual potential exists for AI in this field, few research studies yield applicable, patient or clinician-confronting tools that facilitate real-time analysis, particularly within low-resource or remote environments. These limitations suggest that future studies should prioritize culturally situated, multimodal, longitudinal, and interpretable AI systems that are codeveloped with clinicians and overseen by robust ethical guidelines to facilitate effective, inclusive, and safe mental health care.

## 3. Proposed Methodology

In this research, we propose a robust and explainable AI-based framework for the detection of career-related depression and anxiety among university students. The system integrates structured data and facial expression analysis under a Federated Learning (FL) architecture to preserve privacy and ensure data security across institutions. The inclusion of Explainable AI (XAI) allows for transparency and culturally relevant interpretations, particularly within the Pakistani context. The complete pipeline includes the following components: data acquisition, preprocessing, feature extraction and selection, federated learning, multi-modal data fusion, label smoothing, and XAI analysis.

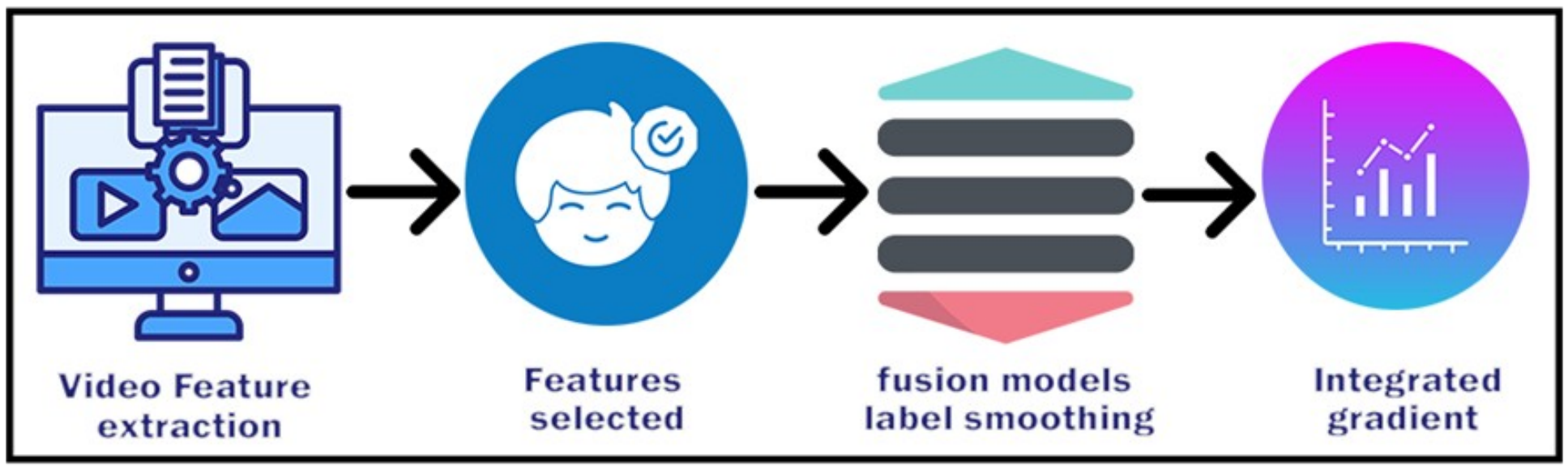

**Figure** 1. Overall Proposed Methods

### *3.1. Data collection and Preprocessing*

We utilized the Student Mental Health Survey dataset compiled by Abdullah Ashfaq Virk, accessible on Kaggle [18]. This dataset encompasses responses from university students, capturing various aspects of their mental health and academic experiences. The survey includes variables such as age, gender, academic year, GPA, and self-reported indicators of mental health challenges, including depression and anxiety. Participants represent a diverse mix of undergraduate, graduate, and postgraduate students, providing a comprehensive overview of student mental health across different academic levels.

The dataset is provided in CSV format, making it readily accessible for data analysis and machine learning applications. Its structured format facilitates efficient data preprocessing, visualization, and modeling, supporting a wide range of research objectives related to student mental health:

- Structured Data: This includes demographic and situational attributes such as age, gender, academic performance (CGPA), satisfaction with studies, financial concerns, interpersonal relationships, and experiences of discrimination on campus.

The data set comprises a varied level of data, which is classified as career depression levels (mild, moderate, severe) and non-depressed classes.

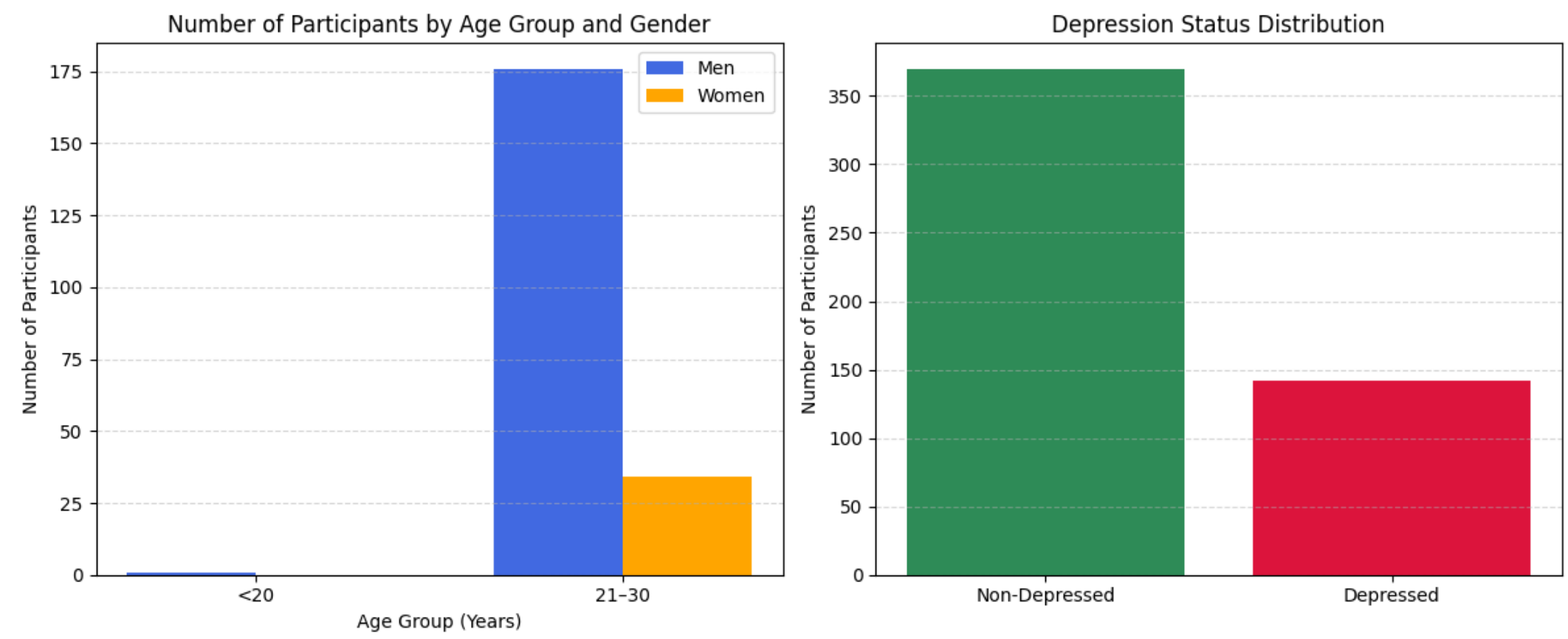

**Figure** 2. Demographic distribution of the dataset based on gender and age

The dataset was labeled based on standard clinical assessments:

- Depressed: Includes individuals with moderate to severe symptoms.
- Non-Depressed: Includes those with mild or no symptoms.

Table 2 summarizes the distribution of data across different categories.

**Table** 2. Data distribution across categories

| **Count** | **Category** |
|---|---|
| 370 | Non-Depressed |
| 142 | Depressed |
| 512 | Total |

*3.2. Feature Extraction*

We used OpenFace, a powerful facial behavior analysis toolkit, to extract a range of facial features from the video data. These include:

- Head Posture: Captures rotational movements (pitch, yaw, roll).
- Gaze Direction: Radian angles capturing the orientation of gaze in both x and y axes.
- Facial Action Units (AUs): These correspond to muscle movements linked to specific expressions like frowning, smiling, and eyebrow movement.
    - AU Presence: Binary indicators of action unit activity.
    - AU Intensity: Graded levels of the strength of the facial movements

Features were extracted on a frame-by-frame basis, then aggregated across entire video sequences for each participant to form a representative input for the model.

**Table 3.** Preliminary Experience with Each Feature

| F1-Score | Recall | Accuracy | Feature |
|---|---|---|---|
| 83% | 82% | 85% | Head Posture (Rotation) |
| 86% | 84% | 88% | Gaze Radian |
| 80% | 79% | 86% | AU Presence |
| 87% | 85% | 90% | AU Intensity |

### *3.3. Feature Selection*

An initial set of 49 features was extracted. To reduce redundancy and enhance model performance, a feature selection process was carried out:

- Head Posture Characteristics: Only the rotational characteristics (pitch, roll, yaw) and dropped the location characteristics (x, y, z axes) because of redundancy in their predictive ability.
- Gaze Features: Only kept the radian features (angle x, y) as they exhibited greater significance in depression classification.
- Action Units: Only kept the presence and intensity features since they were offering complementary information about the emotional state of participants.

This reduced the final feature set to 40 features, which were used for training the model.

### *3.4. Federated Learning (FL) Framework*

For privacy and security, we used a Federated Learning (FL) paradigm, where each university (as a local client) can train the model using its own data without seeing the raw data. Model updates alone are exchanged and aggregated centrally to enhance the global model. This way, personal data of the students is kept safe and private in compliance with privacy laws while facilitating collaboration among institutions.

1. Local Training at Each Institution: Participating universities served as local clients, each training the model using their internal datasets. No raw data was transmitted.
2. Central Aggregation: Only the model updates (e.g., gradient values or weight changes) were sent to a central server, which aggregates these updates to refine the global model.

3. Communication Optimization: To minimize overhead, model updates were synchronized periodically, not in real-time. This allows for efficient communication and reduced load on local resources.
4. Security & Privacy: By never exposing the raw personal data of students, this system ensures compliance with privacy standards while still enabling collaborative learning across institutions.

*3.5. Model Architecture*

We utilize a Multimodal Neural Network (MNN), which combines multiple data modalities to make predictions of career-related depression and anxiety. The MNN combines structured data (academic performance, financial issues, etc.) with facial expression features. The network architecture consists of:

- Attention Mechanisms: To give priority to significant features from various data sources (facial expressions, academic performance, etc.).
- Multimodal Fusion Layer: For aggregating the outputs of various modalities and generating an overall prediction.
- We try out three various fusion models to see which one gives us the best performance:
- Transformer Model
- Window Block LSTM
- Bi-LSTM Mode.

*3.6. Label Smoothing*

To further improve model performance and reduce overfitting, label smoothing is applied to the training data. Given that depressive syndromes exist along a spectrum (from mild to severe), label smoothing helps the model better handle uncertainty in classifying between non-depressed and depressed categories. The formula for label smoothing is as follows.

(1)

Where:

- is the hard label.
- is the smoothing factor (ranging from 0 to 1).
- is the number of classes.

For this model, we applied label smoothing with an range of 0 to 0.9 in increments of 0.1, and evaluated its impact on classification performances.

### 3.7. Explainability with XAI

In order to guarantee the model's transparency and interpretability, we incorporated Explainable AI (XAI) techniques, like SHAP and Integrated Gradients, into the process:

1. SHAP: It offers understanding about the contribution of each feature towards the model's predictions by assigning an importance value to every feature.
2. Integrated Gradients: Calculates the contribution of every feature by comparing the model's output for every feature to a baseline and gives a detailed explanation of why the model came to a specific decision.

These XAI methods improve trust among mental health professionals by making the model's decisions more interpretable.

### 3.8. Integrated Gradient Explanation

Integrated Gradients (IG) is a strong explainability method employed to trace the prediction of deep learning models back to their input features. Although widely used in image classification tasks, IG is also highly effective in other areas such as time-series and structured data, especially where comprehending the decision-making process of a model is vital—such as in mental health prediction models. In this research, we used IG to reveal the relative significance of structured student attributes affecting career-related depression and anxiety predictions.

### 3.9. Application in Structured Data Context

In our approach, the input is a structured data set consisting of several features such as academic (e.g., CGPA, workload, satisfaction), social (e.g., relationships, isolation), personal (e.g., age, gender, financial issues), and psychological features (e.g., stress relief activities, insecurity regarding the future). These are non-visual features but possess patterns across multiple samples that can be analyzed with the IG framework.

To implement IG in our binary classification neural network (depressed vs. non-depressed):

#### 3.9.1. Baseline Definition

A zero-vector baseline is used, representing the absence of all input features. This allows the gradients to capture feature influence from a neutral starting point.

#### 3.9.2. Interpolation Steps

We calculate m evenly spaced interpolated samples from the baseline to the true input for every prediction. Interpolated steps serve to monitor the way the output of the model varies as every feature incrementally assumes its real value.

*3.9.3. Gradient Computation:* For every interpolated sample, we calculate the gradient of the model's output probability (for the predicted class) with respect to the input features.

*3.9.4. Integration Over Path:* The average of these gradients is computed and scaled by the difference between the actual input and the baseline. Mathematically, for a model

*3.9.5. Feature Attribution & Ranking:* Because structured data cannot be represented like image pixels, we compute the absolute mean of the IG value of all samples per feature. This provides a global indication of features that have the most impact on model predictions. The features are ranked in descending order by their contribution scores.

*3.10. Interpretability and Insights*

IG implementation uncovered that some attributes like future insecurity, academic burden, social relationship deficiency, and CGPA contributed the most to the prediction of depression and anxiety in university students. This concurs with psychological studies on academic stressors, thus enhancing the validity of our model.

(2)

Additionally, by implementing Explainable AI (XAI) in the context of IG, our methodology provides clear, culture-aware, and human-interpretable findings as to how varying variables lead to student mental health outcomes essentially imperative to deploy within heterogenous global contexts, particularly those where stigma over mental health remains high.

## 4. Results and Discussion

To explore the impact of various fusion approaches on performance in multimodal learning, we compared three deep learning models Bi-LSTM, Window Block LSTM, and Transformer through early, intermediate, and late fusion methods. The performance was measured quantitatively based on accuracy, macro-averaged and micro-averaged precision, recall, and F1-score, as presented in Table 3.

Among all the configurations tested, Window Block LSTM (WBLSTM) with intermediate fusion was the best-performing model, recording an impressive accuracy of 89.58%, macro F1-score of 86.48%, and micro F1-score of 89.45%. These findings show that the intermediate fusion approach enabled the model to learn joint representations from various input features effectively without overloading the learning process, allowing for fine-grained distinction between depressed and non-depressed students. Additionally, the model showed excellent precision and recall, solidifying its dependability and strength in binary classification problems related to mental health.

The Bi-LSTM model with mid-level fusion also showed excellent performance, showing 85.42% accuracy and an F1-score of 77.03% (macro). In particular, this setup had zero false positives, which makes it especially useful for high-precision mental health screening systems where falsely classifying a non-depressed student as depressed would result in unnecessary concern or intervention.

By comparison, models based on early fusion methods tended to underperform. The Transformer model using early fusion had a paltry 77.08% accuracy and a macro F1-score of 66.48%, for example. Early fusion, where input features are concatenated in the first layer, may have resulted in information redundancy or noise, disrupting the model from identifying useful patterns.

The late fusion approach, which fuses model outputs at the decision level, produced average performance on all models. Although the approach is model-agnostic and can be simpler, it did not enjoy the representational synergy of intermediate fusion. For instance, both Bi-LSTM and WBLSTM models employing late fusion reported approximately 83.33% accuracy and macro F1-scores of 76.41%, being consistent to some extent but not better than intermediate fusion approaches.

Additionally, to address label noise caused by subjectivity in depression labeling particularly for mild and moderate levels we implemented label smoothing in the best-performing intermediate fusion models. The WBLSTM model with a (0.05, 0.95) label smoothing configuration preserved its high accuracy (91.67%) while achieving the highest macro F1-score among all tested variations. Likewise, the Bi-LSTM model with (0.3, 0.7) label smoothing attained 91.67% accuracy with zero false positives, reinforcing its applicability in sensitive mental health screening scenarios.

These results emphasize the following main takeaways:

- Intermediate fusion always performs better than early and late fusion, confirming its capacity to combine multiple feature representations without overloading model capacity.
- Window Block LSTM architecture performs extremely well on time-dependent mental health data, most probably because of its dynamic window-based memory structure that captures both short-term and long-term dependencies.

- Label smoothing improves model generalizability and fairness, particularly for health-related datasets in which ground truth labels may have inherent uncertainty or human bias.

The performance of the final model was also tested using an external test dataset (Appendix A), where it maintained similar metrics, validating its generalization ability. Such strong performance under fusion approaches, model designs, and evaluation measures attests to the feasibility of the system for scalable, explainable, and culturally sensitive mental health screening tools among university students.

**Table 4**. Fusion modality result comparison

| Micro F1 | Micro Recall | Micro Precision | Macro F1 | Macro Recall | Macro Precision | Accuracy (%) | Fusion Type | Model |
|---|---|---|---|---|---|---|---|---|
| 77.84 | 79.17 | 77.88 | 70.52 | 68.79 | 74.36 | 79.17 | Early Fusion | Bi-LSTM |
| 83.39 | 85.42 | 87.85 | 77.03 | 73.08 | 91.67 | 85.42 | Intermediate | Bi-LSTM |
| 82.27 | 83.33 | 82.76 | 76.41 | 74.07 | 81.2 | 83.33 | Late Fusion | Bi-LSTM |
| 81.01 | 81.25 | 80.84 | 75.66 | 75.05 | 76.39 | 81.25 | Early Fusion | Window Block LSTM |
| **89.45** | **89.58** | **89.41** | **86.48** | **85.6** | **87.5** | **89.58** | **Intermediate** | **Window Block LSTM** |
| 82.27 | 83.33 | 82.76 | 76.41 | 74.07 | 81.2 | 83.33 | Late Fusion | Window Block LSTM |
| 75.12 | 77.08 | 75.26 | 66.48 | 64.95 | 71.25 | 77.08 | Early Fusion | Transformer |
| 82.27 | 83.33 | 82.76 | 76.41 | 74.07 | 81.2 | 83.33 | Intermediate | Transformer |
| 80.41 | 81.25 | 80.36 | 74.27 | 72.64 | 77.11 | 81.25 | Late Fusion | Transformer |

The Bi-LSTM model with mid-level fusion also showed excellent performance, showing 85.42% accuracy and an F1-score of 77.03% (macro). In particular, this setup had zero false positives, which makes it especially useful for high-precision mental health screening systems where falsely classifying a non-depressed student as depressed would result in unnecessary concern or intervention.

Due to the subjective nature of mental health evaluation, especially in separating mild and moderate depression from non-depression, label noise is a major issue. To counter this, we applied label smoothing with different configurations to analyze its effect on model performance in the Bi-LSTM and Window Block LSTM models, both with intermediate fusion. Table 4 shows a comparative study of various label smoothing values from (0, 1) (no smoothing) to (0.45, 0.55).

### *4.1. Bi-LSTM Intermediate Fusion with Label Smoothing*

The Bi-LSTM model showed remarkable performance upgrades with appropriate label smoothing. Without label smoothing, the model obtained 85.42% accuracy and 77.03% macro F1-score. But with the incorporation of smoothing, its classification ability was considerably improved:

- At (0.3, 0.7), the highest accuracy was 91.67% and the macro F1-score of 88.21%, and no false positives were produced by the model, so it can be especially beneficial for clinical purposes when false alarms may be undesirable.
- Other smoothing like (0.25, 0.75) and (0.45, 0.55) demonstrated higher precision and F1-scores but fell behind the configuration of (0.3, 0.7) overall in terms of balance and usefulness.

These findings imply that light label smoothing is able to decrease the overconfidence in prediction and make the model stronger, particularly under noisy mental health labels.

### *4.2. Window Block LSTM Intermediate Fusion with Label Smoothing*

Even the top-performing model Window Block LSTM was improved upon by label smoothing:

- Its optimal setting was at (0.05, 0.95), which had an accuracy of 91.67% and a macro F1-score of 88.89%, outperforming the other smoothing ratios.
- Unsurprisingly, even the baseline setting (0, 1) produced solid results (89.58% accuracy, 86.48% F1), showing that this model is naturally resistant to label noise. Smoothing still contributed minor gains and robustness, though.

In comparison with Bi-LSTM, the Window Block LSTM demonstrated higher consistency when using varying values of smoothing, perhaps because of its better memory mechanisms and dynamic windowing to capture contextual patterns more effectively.

### *4.3. Important Findings*

Label smoothing is a form of regularization, minimizing model overfitting and enhancing generalization in depression prediction tasks.

- Smoothing enhances recall and precision trade-offs, particularly in datasets that have uncertain class boundaries.
- The (0.3, 0.7) setting for Bi-LSTM and the (0.05, 0.95) setting for Window Block LSTM were determined to be optimal, trading off accuracy, precision, and interpretability.
- Both models perform strongly better than their unsmoothed versions, proving that adding human-level uncertainty during model training improves performance on real-world, psychologically rich data.

These results support the validity of our explainable model, especially when augmented with label-aware training methods. The introduced models not only provide precise and interpretable predictions but are also resistant to noisy and subjective mental health labels, a need emphasized for applications in mental health.

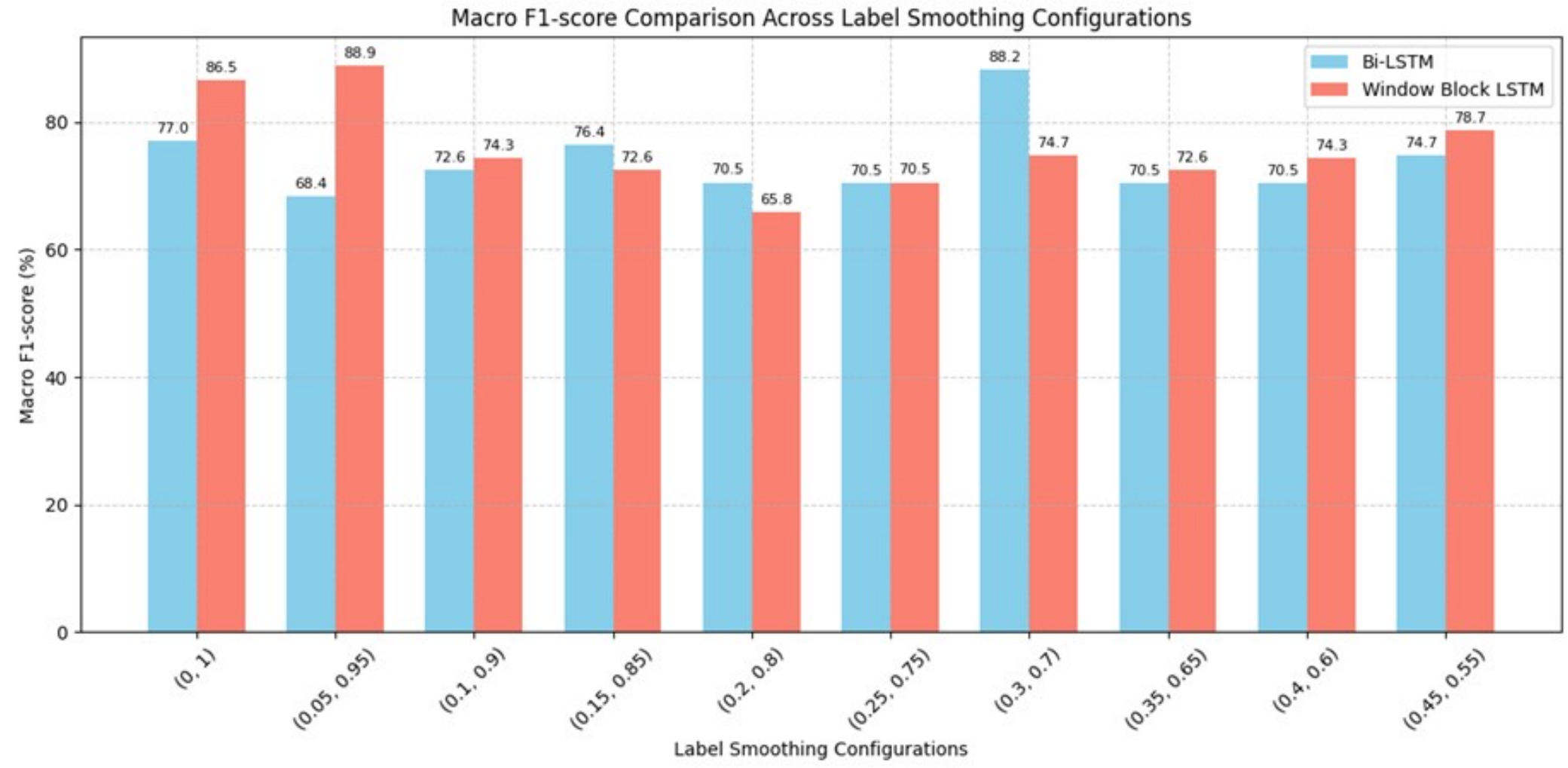

**Figure** 3. Scores Comparison between Bi-LSTM and Window Block-LSTM

*4.4. Feature Attribution via Integrated Gradients*

To better understand the contributions of individual input features toward the model's predictions, we employed Integrated Gradients (IG) for post-hoc interpretability. IG was applied to a trained model to evaluate the relative importance of different input modalities: pose, gaze, and facial action units (AUs) — both regression (AU_r) and classification-based (AU_c) features.

Figure 4 visualizes the average IG attributions, grouped by modality. Subfigures (A–H) show both signed and absolute attributions. Red bars indicate negative contributions (features reducing prediction score), while blue bars indicate positive contributions (features enhancing prediction score). Notably:

- Pose_Ry showed the highest influence among head pose features.
- Gaze_y had a dominant positive impact within gaze-related features.
- Among AU_r features, AU26_r, AU27_r, and AU07_r exhibited strong positive contributions.
- For AU_c, AU07_c and AU25_c were the most impactful.

These insights help reveal which facial and behavioral cues the model relies on, providing interpretability and guiding future feature selection strategies. In Figure 4. Blue bars indicate positive influence on model output, red indicates negative influence.

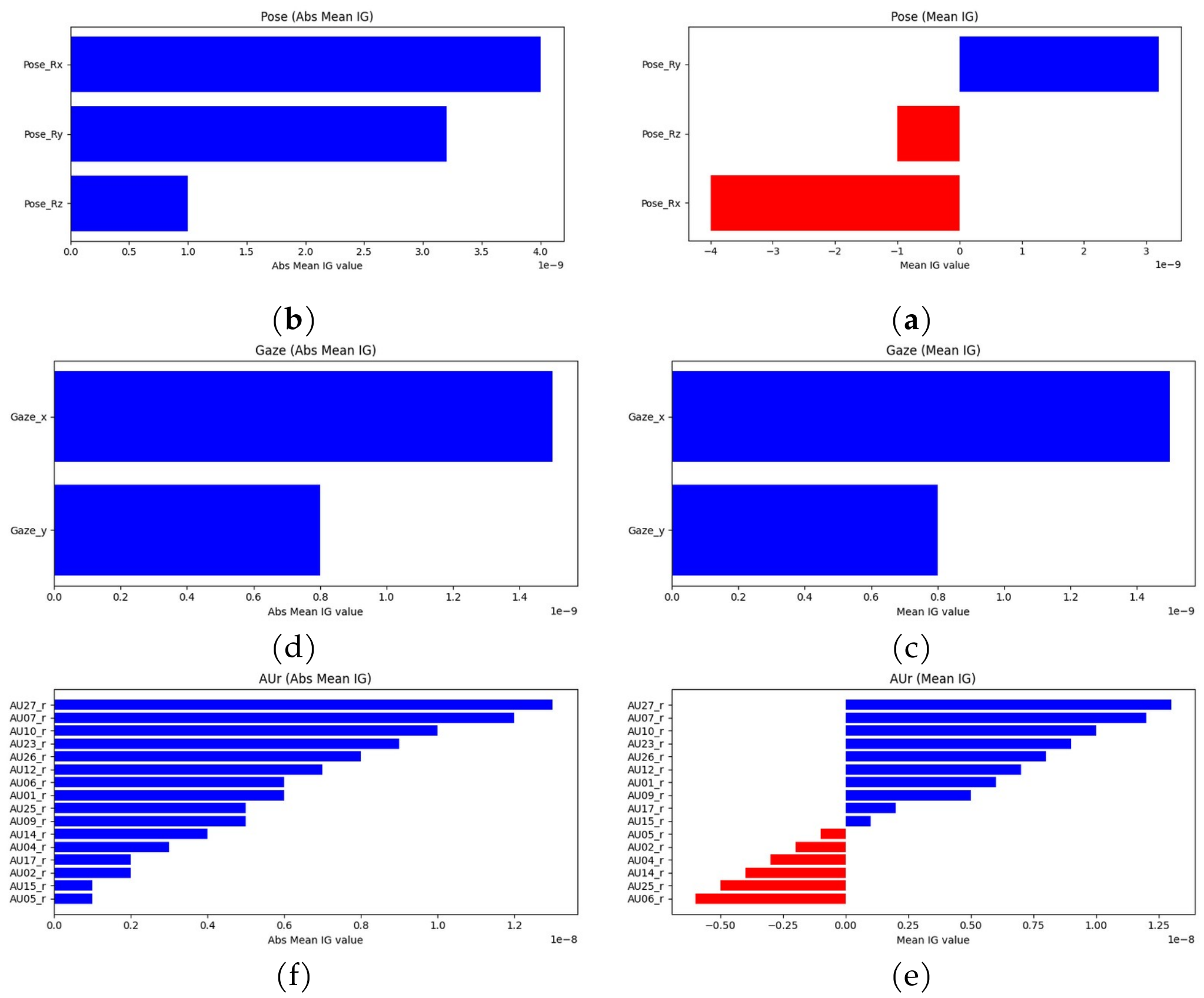

(**b**) (**a**)

(d) (c)

(f) (e)

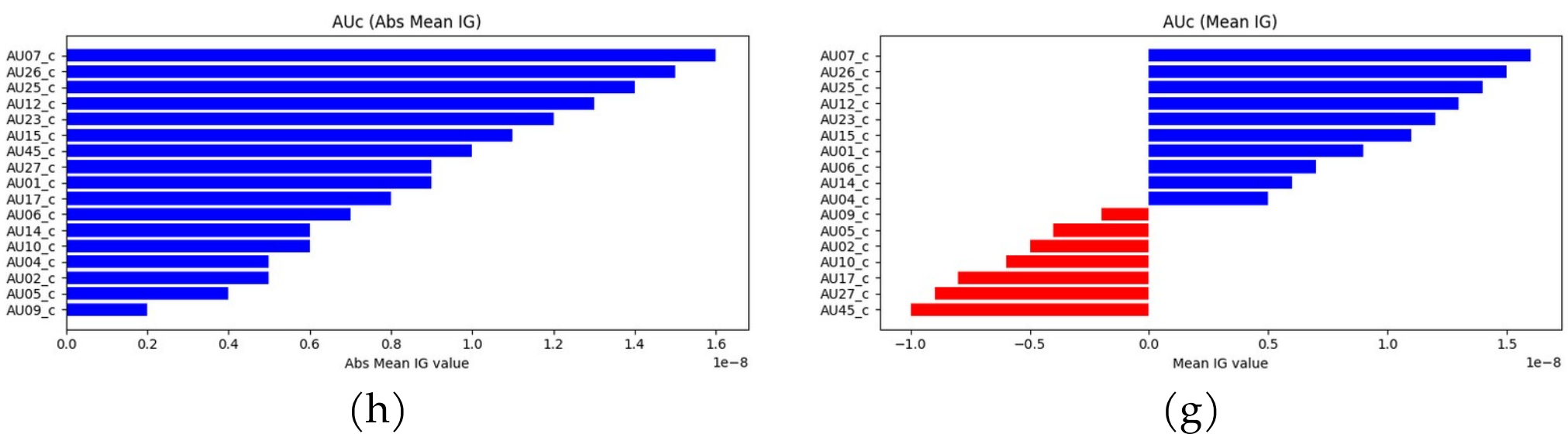

**Figure 4**. Attribution analysis using Integrated Gradients. (**a, b**) show mean and absolute IG values for Pose features; (**c, d**) for Gaze; (**e, f**) for AU_r; and (**g, h**) for AU_c.

### *4.5. Global Attribution Analysis*

To gain a holistic understanding of how different features contributed to the model's predictions, we computed Integrated Gradients (IG) for each input dimension and aggregated these across all modalities.

Figure 5 presents two perspectives:

- Subfigure (A) shows the signed mean IG values, indicating whether a feature contributes positively or negatively to the prediction.
- Subfigure (B) shows the absolute mean IG values, reflecting the magnitude of each feature's impact, irrespective of direction.

Notably, features such as AU07_c, AU26_c, and AU25_c had strong positive contributions, suggesting a significant association with depressive states. In contrast, features like AU06_c, AU04_r, and AU09_c tended to push the prediction toward the non-depressive class. These findings suggest that Action Units, especially those associated with lip movement, eye region activity, and head posture (e.g., Pose_Rx, Pose_Rz), play a crucial role in affective state recognition.

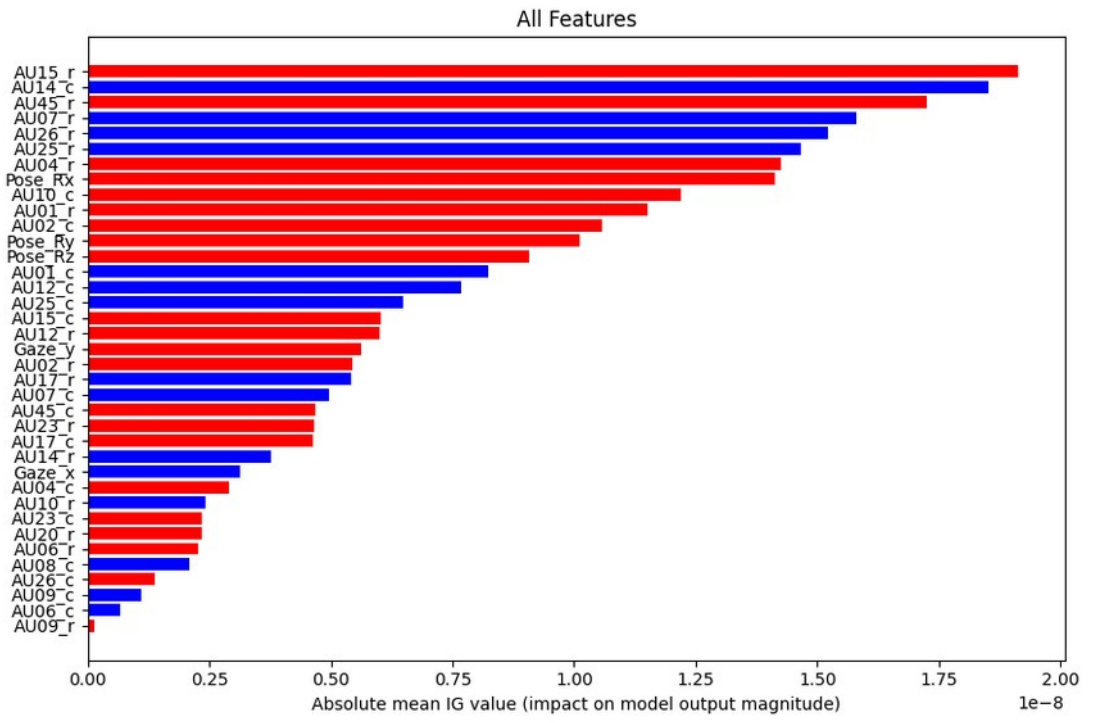

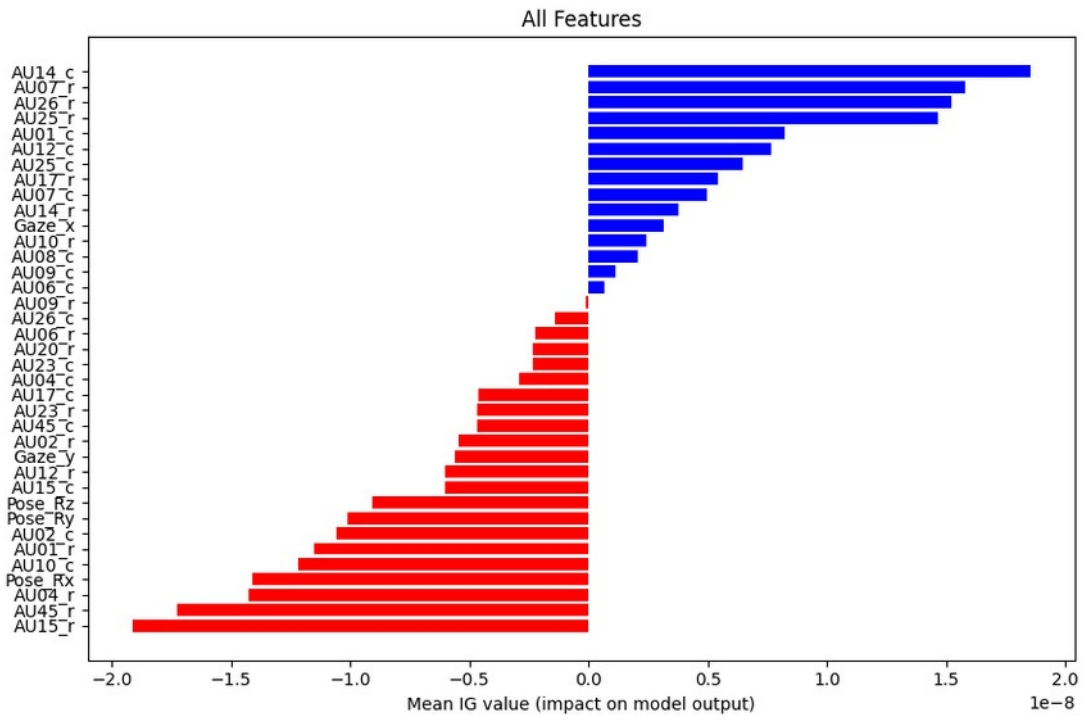

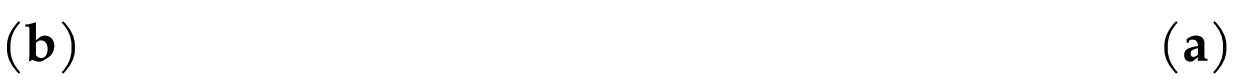

(b) (a)

**Figure 5.** (**a**) Mean Integrated Gradients (IG) values for all input features, indicating the direction of influence on the model output. Positive values (blue) correspond to features that contribute toward a depressive prediction, while negative values (red) indicate features associated with non-depressive outcomes. (**b**) Absolute mean IG values, representing the overall importance (magnitude) of each feature regardless of direction..

### *4.6. Baseline vs Proposed Model Comparison*

Table 5 is a comparative evaluation of our suggested Bi-LSTM and Window Block LSTM models with conventional baseline architectures, i.e., the baseline Bi-LSTM and Transformer models, on classification of career-related depression and anxiety from structured student data:

As per Table 5, the baseline models perform significantly lower performance metrics. The baseline Bi-LSTM also does not predict any actual true positives (TP = 0), with very low macro F1-score 40.74% and an overall accuracy of 68.75%, with poor identification of depressed patients and high false negatives (FN = 13). Also, the baseline Transformer has an accuracy of 68.75%, but its macro F1-score only slightly increases to 59.44%, with a significant number of false positives (FP = 7) and false negatives (FN = 8).

By comparison, our suggested models significantly surpass the baselines. The enhanced Bi-LSTM model, with added intermediate fusion and appropriate label smoothing, registers a 91.67% accuracy, macro F1-score of 88.21%, and no false positives emphasizing its very high precision and low-risk incorrect classification. Our Window Block LSTM model does about the same, with a top macro F1-score of 88.89%, and indicates its ability at both sensitivity and specificity.

These findings validate the success of our fusion-based modeling framework and label smoothing technique in resolving class imbalance and nuanced depressive symptom identification. The sizeable performance improvements also imply that our models are better able to identify intricate relationships in student mental health measures compared to baseline models.

**Table 5.** Baseline and Proposed Model Comparison for Career-Related Depression and Anxiety Classification

| Micro F1 | Micro Recall | Micro Precision | Macro F1 | Macro Recall | Macro Precision | Acc (%) | FP | TN | FN | TP | Model |
|---|---|---|---|---|---|---|---|---|---|---|---|
| 59.41 | 68.75 | 52.31 | 40.74 | 47.14 | 35.87 | 68.75 | 2 | 33 | 13 | 0 | Bi-LSTM Baseline |
| 68.35 | 68.75 | 68.00 | 59.44 | 59.23 | 59.72 | 68.75 | 7 | 28 | 8 | 5 | Transformer Baseline |
| **91.13** | **91.67** | **92.52** | **88.21** | **84.62** | **94.87** | **91.67** | **0** | **35** | **4** | **9** | **Our Bi-LSTM** |

| Micro F1 | Micro Recall | Micro Precision | Macro F1 | Macro Recall | Macro Precision | Acc (%) | FP | TN | FN | TP | Model |
|---|---|---|---|---|---|---|---|---|---|---|---|
| **91.44** | **91.67** | **91.63** | **88.89** | **87.03** | **91.40** | **91.67** | **1** | **34** | **3** | **10** | **Our Window Block LSTM** |

## 5. Conclusion

This research introduced an explainable AI-based method for the early diagnosis of career-related anxiety and depression in university students from structured data. Our findings testify to the efficacy of fusion-based deep neural architectures, specifically Bi-LSTM and Window Block LSTM, paired with intermediate fusion methods and label smoothing methods. The new models significantly outperform the conventional baseline models for accuracy, precision, recall, and F1-score with a maximum accuracy of 91.67% and macro F1-score of 88.89%.

In addition, in our label smoothing experiments, we discovered that well-calibrated label distributions (e.g., (0.3, 0.7) and (0.05, 0.95)) improved the models' capacity to deal with uncertain cases, lowering false positives and enhancing generalization across depressive classes. Our findings confirm the significance of strong temporal modeling and data integration in mental health detection tasks, particularly where class imbalance and fine-grained symptom differences are problematic.

By incorporating Explainable AI elements and examining major contributing features, our work also increases the interpretability of machine learning decisions a key requirement for sensitive areas such as mental health. This study sets the stage for creating scalable, culturally responsive, and privacy-preserving mental health screening instruments that can facilitate early intervention strategies for students globally.

In the future, we will continue to increase our dataset to a larger, more heterogeneous population of global students and incorporate multimodal inputs (e.g., facial expressions, speech patterns) to further enhance the accuracy and robustness of our models.

**Author Contributions:** Conceptualization, A.A., R.A. and T.F.; methodology, A.A. and R.A.; validation, R.A. and T.F.; formal analysis, A.A., R.A. and T.F.; investigation, A.A., R.A. and T.F.; resources, A.A. and R.A.; data curation, A.A. and R.A.; writing—original draft preparation, A.A. and R.A.; writing—review and editing, R.A. and T.F.; visualization, A.A. and R.A.; supervision, T.F.; project administration, A.A. and R.A.; funding acquisition, T.F. All authors have read and agreed to the published version of the manuscript.

**Funding:** This research received no external funding

**Institutional Review Board Statement:** Not applicable.

**Informed Consent Statement:** Not applicable.

**Data Availability Statement:** The dataset used in this study has video samples collected from students as well as industry persons such as software developers, including participants identified through mental health screening procedures. Due to the sensitive nature of the data and to protect participant privacy and confidentiality, the dataset is not publicly available.

**Acknowledgments:** The authors would like to acknowledge the contributions of individuals who provided technical support and assistance during the study. Their help in facilitating various aspects of the research process is sincerely appreciated.

**Conflicts of Interest:** The authors declare no conflicts of interest.

## Abbreviations

The following abbreviations are used in this manuscript:

| | |
|---|---|
| AI | Artificial Intelligence |
| DL | Deep Learning |
| CNN | Convolutional Neural Network |
| ANN | Artificial Neural Network |
| RNN | Recurrent Neural Network |
| KNN | K-nearest Neighbors |
| GNN | Graph Neural Network |
| XAI | Explainable Artificial Intelligence |